# Natural Language Generation in Healthcare: A Review of Methods and Applications

**Authors:** Mengxian Lyu[1], Xiaohan Li[1], Ziyi Chen[1], Jinqian Pan[1], Cheng Peng[1], Sankalp Talankar[1], Yonghui Wu[1,2 *]

**Affiliations:**

[1]Department of Health Outcomes and Biomedical Informatics, College of Medicine, University of Florida, Gainesville, Florida, USA.

[2]Preston A. Wells, Jr. Center for Brain Tumor Therapy, Lillian S. Wells Department of Neurosurgery, University of Florida, Gainesville, Florida, USA.

*Corresponding author

Yonghui Wu, PhD

The Malachowsky Data Science & Information Technology Building

1889 Museum Rd, 7th Floor, Suite 7000, Gainesville, FL 32611

Phone: 352-294-8436

Email: yonghui.wu@ufl.edu

**Abstract**

Natural language generation (NLG) is the key technology to achieve generative artificial intelligence (AI). With the breakthroughs in large language models (LLMs), NLG has been widely used in various medical applications, demonstrating the potential to enhance clinical workflows, support clinical decision-making, and improve clinical documentation. Heterogeneous and diverse medical data modalities, such as medical text, images, and knowledge bases, are utilized in NLG. Researchers have proposed many generative models and applied them in a number of healthcare applications. There is a need for a comprehensive review of NLG methods and applications in the medical domain. In this study, we systematically reviewed 113 scientific publications from a total of 3,988 NLG-related articles identified using a literature search, focusing on data modality, model architecture, clinical applications, and evaluation methods. Following PRISMA (Preferred Reporting Items for Systematic reviews and Meta-Analyses) guidelines, we categorize key methods, identify clinical applications, and assess their capabilities, limitations, and emerging challenges. This timely review covers the key NLG technologies and medical applications and provides valuable insights for future studies to leverage NLG to transform medical discovery and healthcare.

## Introduction

Natural language generation (NLG) focuses on developing algorithms that generate coherent and contextually relevant text from a form of input[1]. Early-stage NLG systems relied heavily on predefined templates and rules[2] with limited generation ability. Recent breakthroughs in transformer-based large language models[3–5] (LLMs) have transformed NLG[6] into a key technology to achieve generative artificial intelligence (AI). LLM-based NLG models[7–9] have demonstrated powerful generative ability to support medical discovery and healthcare in many ways that have not been witnessed in previous generations of AI.

Historically, many algorithms have been proposed to achieve NLG, from early-stage statistical machine learning models[10], Long-Short Term Memory (LSTM)-based sequence-to-sequence models[11], to recent transformer-based LLMs[12]. Transformers are a specific type of deep neural network architecture that is composed of an encoder component and a decoder component[13]. LLM-based NLG systems use the encoder to encode an input sequence into vectors, potentially capturing latent semantics, and use the decoder to decode the vectors into human languages. LLMs can be categorized into three types: (1) encoder-based LLMs (e.g., BERT[14]), which are implemented using the encoder module of the transformer; (2) decoder-based LLMs (e.g., GPT[15]), which are implemented using the decoder module of the transformer; and (3) encoder-decoder LLMs (e.g., T5[5]), which are implemented using both the encoder and decoder components. Among these, decoder-based and encoder-decoder LLMs are widely used to achieve generative AI, known as generative LLMs.

In the medical domain, LLMs have demonstrated potential in supporting a wide range of medical discovery and healthcare applications, including extracting patients' information from electronic health records (EHRs)[16,17], differentiating diagnoses for clinical decision support[18,19], answering healthcare-related questions[16], and facilitating documentation of patient information in EHR systems[20]. There has been an increasing interest in using NLG for conversational and generative AI that was not well-performed in the previous generations of AI models[10]. As their multi-modality nature, NLG systems[21–23] utilize different modalities of medical data, such as structured EHR elements in database tables, narrative clinical text, medical images, and videos. NLG systems to generate text from non-linguistic input are often referred to as "data-to-text" generation, and NLG systems to generate text from text input are referred to as "text-to-text" generation. Both generation tasks can be referred to as "text generation". Recent studies suggest that LLM-based NLG systems can assist in documenting patients' reports[24,25] with improved efficiency and quality[26,27], thereby contributing to more streamlined clinical workflows and enhanced healthcare delivery[28]. These emerging use cases highlight the potential of generative AI in supporting real-world clinical practice[29,30].

There is a need for a comprehensive review of NLG focusing on the key breakthroughs of generative AI in the medical domain. Existing review articles mainly focused on traditional natural language processing (NLP) tasks, such as information extraction[31], or narrowly focused on "image-to-text" generation[32], which could not capture the multimodalities of medical data and lacked focus on text generation and generative AI. To address this gap, this literature review provides an overview of recent advances in NLG with a focus on generative LLM methods and applications in the medical domain. Following the PRISMA guidelines, we examine several key applications of NLG, including synthetic clinical text generation, automated clinical documentation, medical summarization, radiology report generation, and medical dialogue systems. This review article aims to answer the following questions: (1) What are the methods used by NLG systems in the medical domain? (2) What are the evaluation metrics for NLG? (3) What are the applications of NLP in the medical domain? (4) What are the capabilities, challenges, limitations, and future directions and opportunities for NLG in the medical domain? To the best of our knowledge, this is the first study to systematically review NLG with multiple modalities of medical data and across various healthcare applications.

## Results

### Study Selection and Characterization

**Fig. 1** is a PRISMA flow diagram showing the study selection process. A total of 3,988 articles published from 2018 to 2024 were identified through database searches using keywords (see **Table 3**). After removing 761 duplicates, 3,227 unique articles remained. The title and abstract screening excluded 2,903 articles, with 324 articles remaining. The 324 remaining articles were further assessed by a "full text review", where 211 studies not focused on healthcare settings or lacking concrete NLG methods were excluded. A total of 113 studies were included in this review, where 76 (67.26%) were conference papers and 37 (32.74%) were journal publications. Detailed characteristics of each study—including model architecture, data sources availability, evaluation methods, and application domain—are summarized in **Supplementary Table S1**.

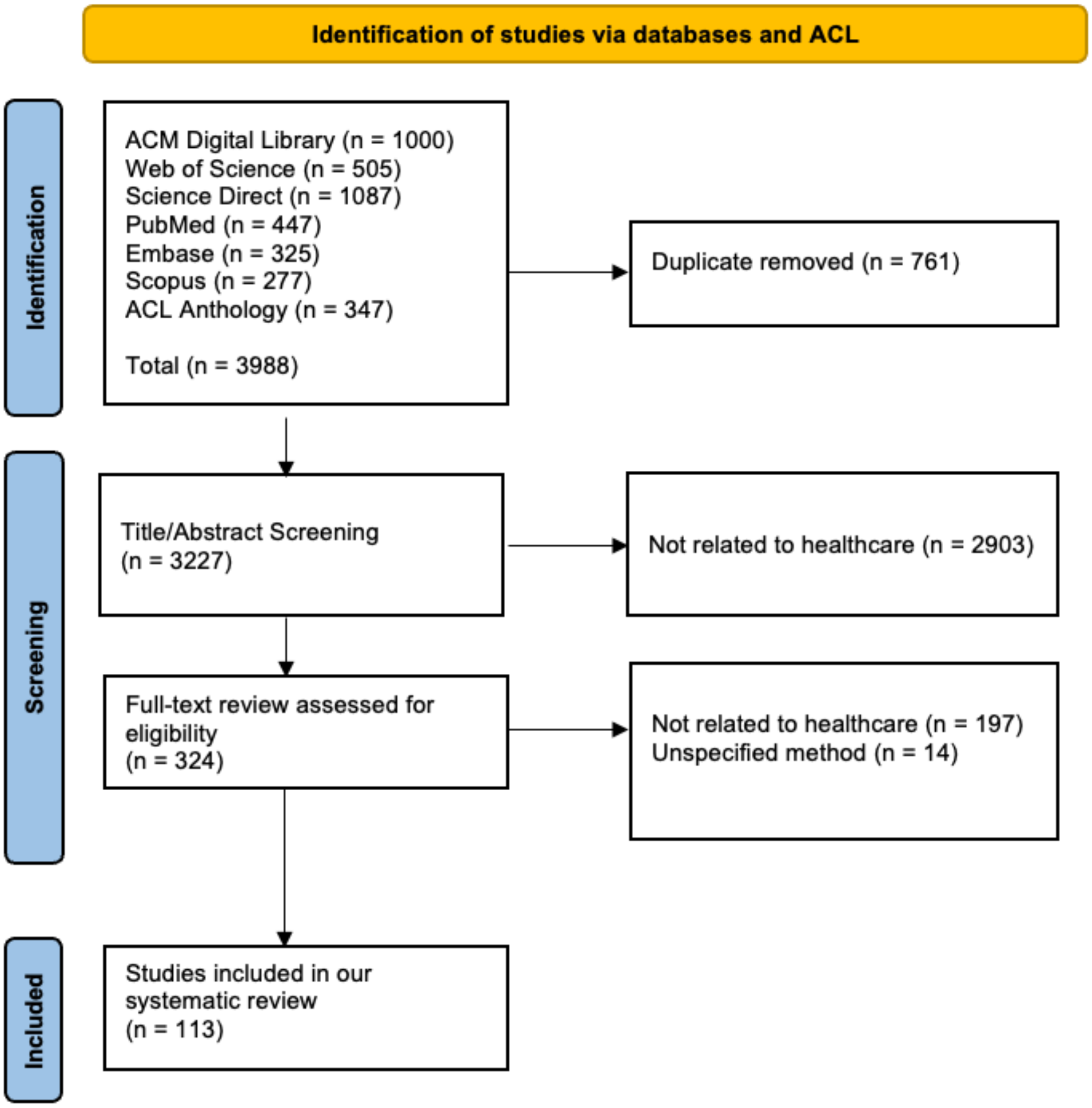

**Fig. 1** PRISMA flow diagram of included studies in this systematic review.

**Fig. 2** summarizes the number of publications related to NLG in the medical domain for the past seven years. The number of publications grew from just 2 in 2018 to 40 in 2024. In terms of geographic distribution, Asia contributed the largest share of studies (50; 44.25%), followed by North America (41; 36.28%) and Europe (16; 14.16%), showing a wide international engagement. Notably, the United States accounted for the largest proportion of studies, contributing 40 publications (35.40%).

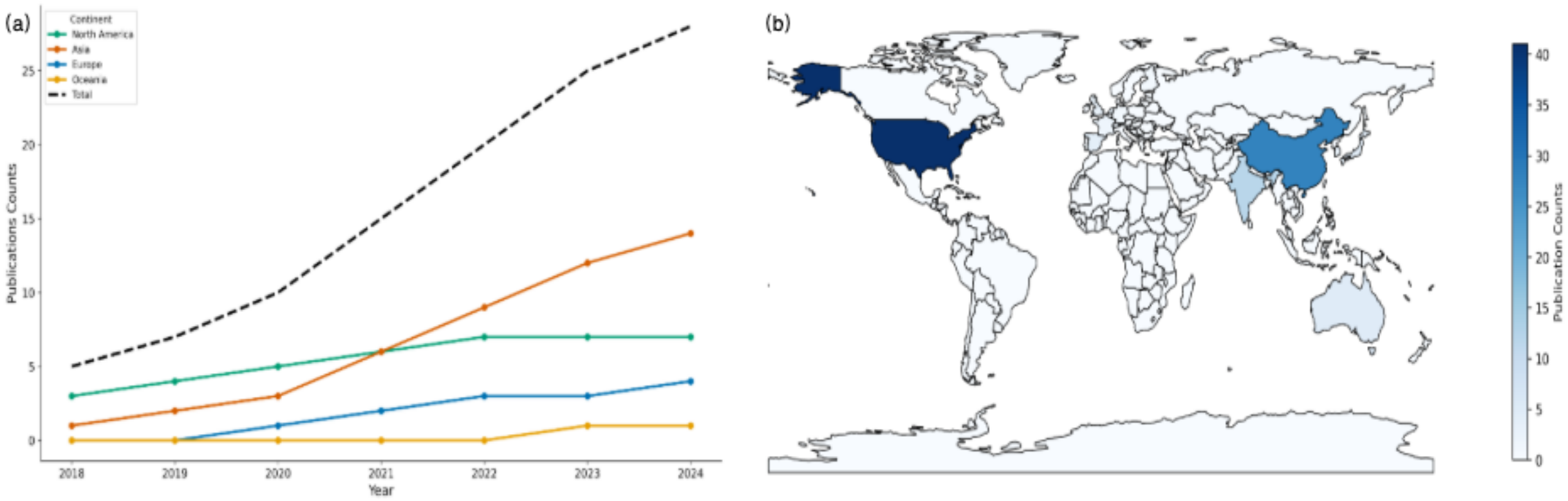

**Fig. 2 Publications on natural language generation in the medical domain by region from 2018 to 2024.** (a) Temporal trends of publication counts by continent, showing the annual number of publications for North America, Asia, Europe, Oceania, and the global total (Total); (b) Geographic distribution of publication counts by country.

## Medical data modalities and text generation applications

Per our review of the 113 included studies, NLG systems in the medical domain mainly use narrative text, structured EHRs, existing medical knowledge bases, and medical images. According to the input modalities, the 113 studies can be grouped into three categories: (1) text-to-text generation, where the input is purely text, (2) image-to-text generation, where the input is medical images, and (3) multimodal-to-text generation, where more than one modality is used. For each category, we examined the overall architecture, the encoder component, which is responsible for transforming input data into vector representations, and the decoder component, which is responsible for converting the vectors into natural language text. **Fig. 3** shows an overview of the medical data modalities and NLG methods summarized from the 113 studies.

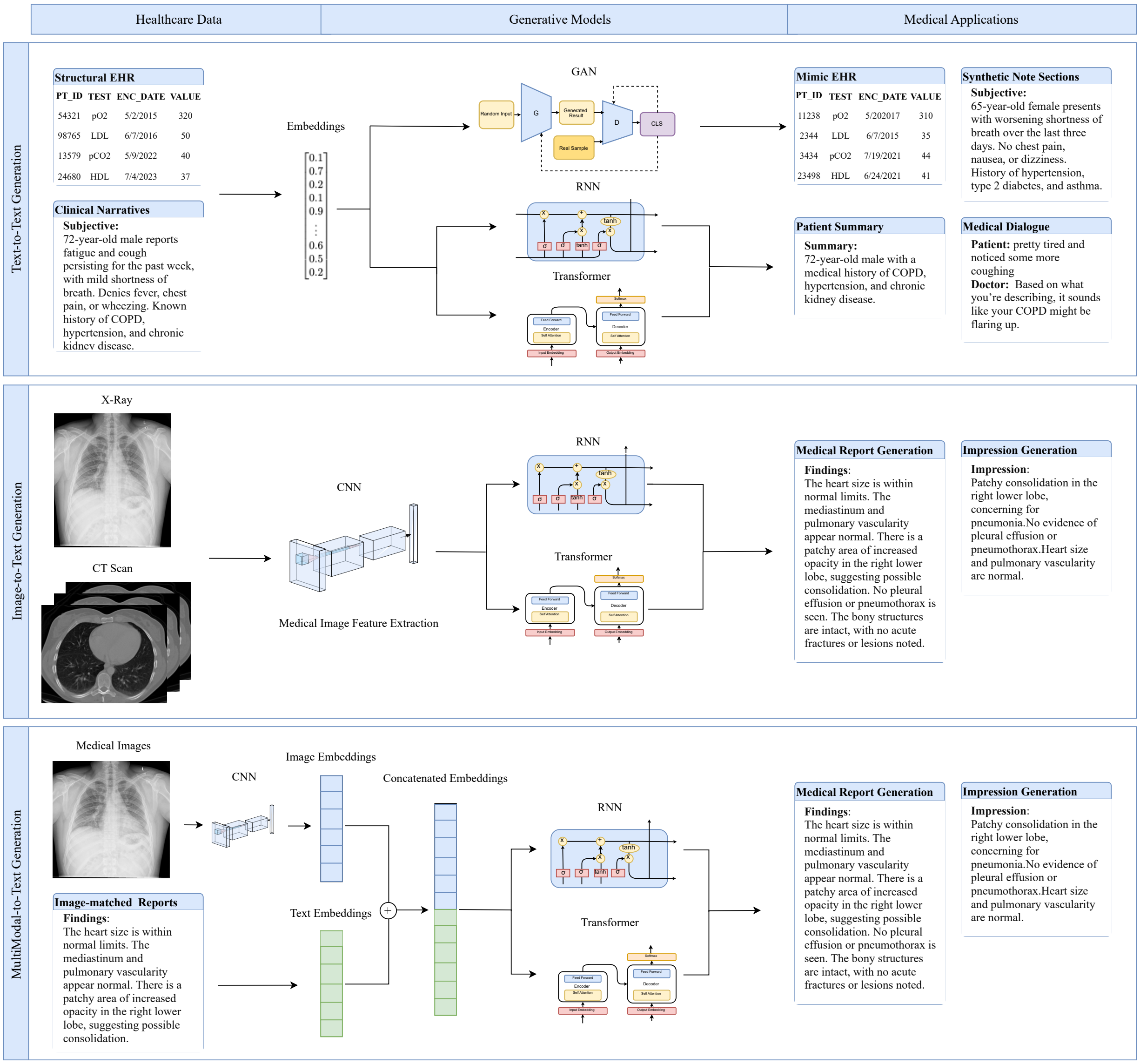

**Fig. 3** An overview of medical data modalities and natural language generation methods. EHR: electronic health record; CNN: convolutional neural network; RNN: recurrent neural network.

## What are the methods used by NLG systems in the medical domain?

**Table 1** aggregates and summarizes the model architectures used by the 113 reviewed studies. There are three mainstream model architectures for NLG, including: (1) encoder-decoder models, (2) decoder-only models, and (3) generative adversarial networks (GANs). Transformer-based models dominate across all modalities. In text-to-text generation, 61.64% (45/73) of studies used encoder-decoder transformer models, and 27.40% (20/73) of studies used decoder-only transformer models. In image-to-text generation, hybrid models combining convolutional neural networks (CNNs) with transformer decoders— CNN+Transformer architectures—were particularly prevalent (~8%), leveraging CNNs for feature extraction from medical images and transformers for text generation. In multimodal-to-text generation, transformer-based architectures were also widely used to integrate heterogeneous data sources such as medical images and structured clinical inputs.

**Table 1.** Detailed model architectures and prevalence across data modalities.

| Modality | Model Architecture | | | Prevalence |
|---|---|---|---|---|
| | Overall Architecture | Encoder | Decoder | |
| Text-to-Text | GAN | - | LSTM | 1 (0.88%) |
| | Encoder-Decoder | FNN | LSTM | 1 (0.88%) |
| | | RNN | RNN | 1 (0.88%) |
| | | GRU | GRU | 3 (2.65%) |
| | | LSTM | LSTM | 4 (3.54%) |
| | | LSTM | Transformer | 1 (0.88%) |
| | | Transformer | Transformer | 42 (37.17%) |
| | Decoder-Only | - | LSTM | 1 (0.88%) |
| | | - | Transformer | 20 (17.70%) |
| Image-to-Text | Encoder-Decoder | CNN | LSTM | 7 (6.19%) |
| | | CNN+LSTM | LSTM | 1 (0.88%) |
| | | CNN +GAN | Transformer | 1 (0.88%) |
| | | CNN+Transformer | Transformer | 9 (7.96%) |
| | | Transformer | LSTM | 1 (0.88%) |
| | | Transformer | Transformer | 3 (2.65%) |
| Multimodal-to-Text | Encoder-Decoder | CNN+GRU | GRU | 1 (0.88%) |
| | | CNN | Transformer | 1 (0.88%) |
| | | CNN+Transformer | Transformer | 9 (7.96%) |
| | | Transformer | Transformer | 6 (5.31%) |

**Text-to-text generation** refers to the task of generating narrative text from textual input, including structured data (e.g., diagnosis codes, medications), semi-structured data (e.g., templates or tabular EHR entries), and unstructured free text (e.g., clinical notes, medical reports). Early studies of text-to-text generation widely used recurrent neural networks (RNNs), particularly gated recurrent unit (GRU) and LSTM. S. Lee *et al.*[25] used a feedforward neural network (FNN) as the encoder and an LSTM as the decoder to generate clinical note sections from structured EHR codes; Melamud & Shivade [33] used a standard LSTM to generate privacy-preserving clinical notes. A small set of studies used GAN for text generation. For example, Guan *et al.*[24] proposed the Medical Text Generative Adversarial Network (mtGAN) to generate synthetic text using a conditional GAN framework with disease features as inputs. Some studies also explored "attention" mechanisms to focus models on clinically relevant content and demonstrated improved quality. For example, J Kurisinkel & Chen *et al.*[34] combined a GRU-based encoder and decoder with attention mechanisms to improve the generation of narrative clinical note sections from structured EHR data.

Most recent studies applied transformer models and demonstrated superior performance in capturing long-distance dependencies and facilitating large-scale pretraining. Both Encoder-Decoder architectures[35] and Decoder-based[7] transformers have been adopted. The pretrain-finetune strategy and various domain-specific adaptations remarkably enhanced the quality and relevance of text generation. Later, advanced techniques such as in-context learning and multi-task instruction tuning further advanced text generation[9,36] in the medical domain.

**Image-to-text generation** refers to the task of generating text reports from medical images, such as chest X-rays, CT scans, pathology images, or other types of medical images. Early studies widely used CNN–RNN architectures, where CNNs were used to extract visual features and RNNs were used as decoders to generate text. Huang *et al.*[37] proposed a hierarchical model that incorporates CNNs with multi-attention mechanisms to enhance image feature representation and combines an LSTM layer to generate chest X-ray reports. Beddiar *et al.*[38] used a pre-trained CNN model to extract visual features and semantic features, which were then forwarded to an LSTM model for text generation. Similar to text-to-text generation, most recent studies applied transformers as decoders, but CNNs are still widely used in the encoder component. For example, Miura *et al.*[39] combined CNNs with a memory-augmented transformer model to generate clinical findings from medical images. Later, vision transformers (ViTs)—which extend transformer architectures to process images directly— demonstrated promising results. For example, Q. Li *et al.*[40] proposed MedEPT, a parameter-efficient model for medical report generation using a pre-trained ViT, and achieved better performance with fewer parameters and shorter training time.

**Multimodal-to-text generation** refers to text generation from multimodal data sources, often combinations of more than one modality from medical images, medical text, and external knowledge graphs. Early studies typically extended the CNN–RNN architecture by incorporating visual and structured features through early or late fusion strategies. For example, Edelbrock *et al.*[41] proposed a model that integrated visual features from X-ray images into an RNN-based encoder–decoder framework to initialize hidden states and modulate encoder outputs and target embeddings, demonstrating improved performance on summarization tasks. More recent studies adopted multimodal Transformer-based architectures. For example, Dalla Serra *et al.*[42] developed a two-step pipeline utilizing multimodal Transformers to first extract clinically relevant triples (Entity1, Relation, Entity2) from radiology images and then apply another Transformer model to generate radiology reports. By controlling text generation using extracted triples and the original images, the proposed method improved both stylistic accuracy and clinical relevance.

## What are the evaluation metrics for NLG?

Text generation tasks are typically evaluated by comparing the generated text with gold-standard references using automatic machine-based and/or human-based evaluation metrics. **Table 2** summarizes the evaluation metrics and their prevalence per our review of the 113 articles.

**Table 2.** Evaluation metrics for text generation.

| Metric Category | Metric | Counts |
|---|---|---|
| Automatic Evaluation | ROUGE[43] | 74 |
| | BLEU[44] | 60 |
| | METEOR[45] | 31 |
| | BERTScore[46] | 20 |
| | BLEURT[47] | 6 |
| Human Evaluation | Likert Scale[48] | 36 |
| | Turing Test | 2 |

For automatic evaluation metrics, the ROUGE and BLEU scores are the most prevalent metrics used by 74 and 60 studies, respectively. Embedding-based methods, such as BERTScore, are also popular (by 20 studies). For human evaluation metrics, the Likert Scale is the most prevalent metric used by 36 studies.

**Automatic Evaluation Metrics**

Automatic evaluation metrics compare string overlap, content overlap, string distance, or lexical diversity between the generated text and gold-standard reference text to calculate various quantitative scores. Widely used automatic evaluation metrics include:

**ROUGE**[43] (Recall-Oriented Understudy for Gisting Evaluation) is the most widely used metric (74/113, 65.49%) reviewed in this study. ROUGE focuses on recall by assessing how many surface-level units, such as N-grams, longest common subsequences, or skip-bigrams from the reference text are replicated by the generated text.

**BLEU**[44] (Bilingual Evaluation Understudy) focuses on N-gram generation accuracy. Per our review, some studies (60/113, 53.10%) exclusively use the BLEU or ROUGE. For example, studies[38,49,50] only used BLEU, while others[51–61] only used ROUGE.

**METEOR**[45] (Metric for Evaluation of Translation with Explicit Ordering) is a special metric that considers precision, recall, and alignment. METEOR considers synonymy and stemming, allowing for a more nuanced evaluation to ensure a comprehensive evaluation of text generation.

### Embedding-based Metrics

Embedding-based metrics such as BERTScore[46] and BLEURT[47] calculate similarity scores by leveraging contextual embeddings from pretrained models (e.g., BERT) instead of surface-level text overlap.

**BERTScore**[46] utilizes embeddings from the BERT[14] model to measure precision, recall, and F1 scores between generated and the ground-truth reference texts. A variation metric, **BLEURT**[47], further fine-tuned BERT on human judgments to focus on semantic similarity and contextual alignment. Both metrics have been utilized in our reviewed studies[62–65].

### Human Evaluation

Human evaluation asks domain experts to evaluate the quality of generated text according to predefined aspects.[66] The Likert scale[48] is widely used in human evaluation, where evaluators rate various aspects of the generated text, such as fluency, coherence, and relevance, using a predefined scale such as 1-5. Many studies[39,52,54,58,62,64,67–70] in the medical domain widely use human evaluation. Pairwise comparison is another widely used human evaluation metric, where domain experts are asked to compare pairs of generated text to determine which one is better according to predefined criteria. For instance, Peng et al.[7] conducted a Turing Test on model-generated clinical notes and human-written notes in terms of readability and clinical relevance.

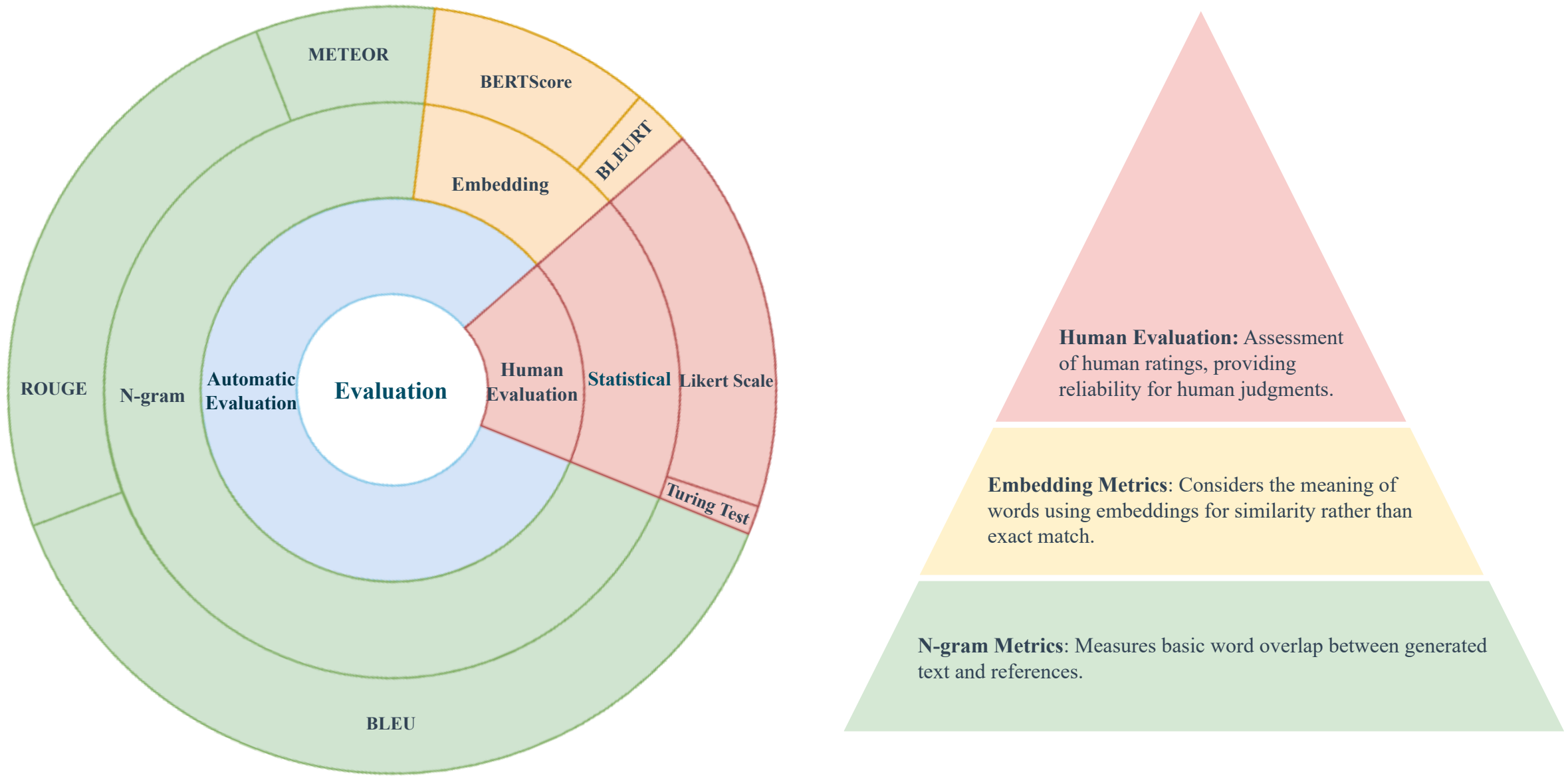

**Fig. 5.** Hierarchical structure of evaluation metrics.

**Fig. 5** presents a hierarchical structure of the evaluation metrics. Human evaluation provides a comprehensive assessment considering nuanced linguistic measures such as fluency, coherence, factual correctness, and clinical relevance—criteria often beyond the reach of automated metrics. However, human evaluation is time and cost-consuming. Embedding-based metrics such as BERTScore and BLEURT measure similarity using embeddings of generated and reference texts, which capture semantic similarities based on embeddings without the requirement of human experts. N-gram-based metrics, including ROUGE, BLEU, and METEOR, calculate similarity using surface-level lexical overlap. These metrics are not able to account for the linguistic variations, as there are many surface-level text variations for the same meaning. Nevertheless, these metrics are widely used as they are easy to calculate.

## Image-to-Text Related Metrics

Image-to-text studies typically use the following domain-specific automatic metrics.

**CIDEr**[71] (Consensus-based Image Description Evaluation) is an automatic metric for evaluating the quality of generated image descriptions compared to human-written references using n-grams and Term Frequency-Inverse Document Frequency (TF-IDF) weighting. CIDEr is widely used in image-to-text generation evaluation in the included studies[37,39,40,72–76].

**Clinical Efficacy** (CE) metrics are designed to evaluate the relevance and accuracy of generated text, focusing on clinically relevant text. For example, one of the metrics [72] uses the CheXpert labeler proposed by Irvin *et al.*[77] to extract 14 types of medical concepts and measures how well the generated text aligns with the reference text on information of clinical importance. The CheXpert-based metrics highly depend on the MIMIC-CXR dataset, limiting their generalizability to other datasets and domains.

**What clinical applications can benefit from NLG?**

Based on our review, we categorize NLG applications in the medical domain into four primary domains, as shown in **Figure 6**:

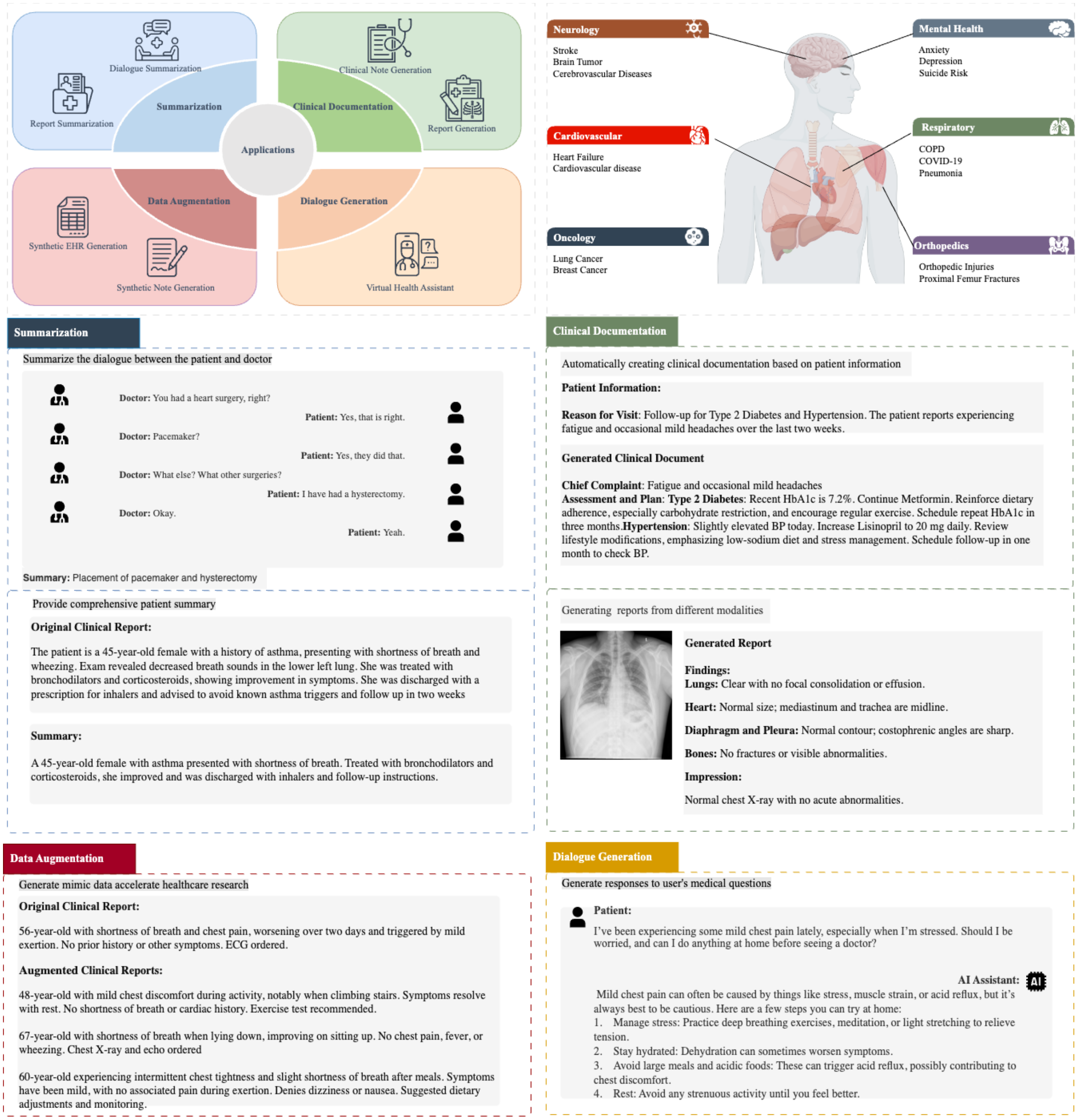

**Fig. 6** Major applications for text generation in the medical domain. (1) **Summarization**, where models condense lengthy medical text into summaries without altering key information and main idea; (2) **Clinical documentation**, including the automatic drafting of clinical notes and diagnostic reports from text-based inputs or imaging data; (3) **Dialogue generation**, which enables contextually relevant responses in healthcare interactions and counseling settings; and (4) **Data augmentation**, where synthetic texts are generated to support healthcare research.

## Clinical Summarization

Clinical summarization aims to transform the original long text into concise summaries while capturing key information[78]. One of the primary applications is radiology report summarization, which focuses on the summarization of the IMPRESSION section and the FINDINGS section in radiology reports. [51,79–81] Another focus is the summarization of clinician–patient encounters. Joshi et al.[67] used text generation to generate patient summaries from telemedicine interactions. Two recent studies[64,82] focused on the summarization of patient problems from progress notes. Krishna et al.[52] applied text generation models to summarize doctor-patient encounters into Subjective, Objective, Assessment, and Plan（SOAP) format. The MEDIQA-Chat 2023[83] shared task provided two dialogue-based summarization benchmarks. Top-performing approaches[60,61,84–86] achieved high scores across standard metrics, including ROUGE, BLEURT, and BERTScore. Recent studies have explored the summarization of discharge summaries. Recent systems[87–89] employed Parameter-efficient fine-tuning (PEFT) and hybrid extractive-abstractive strategies. For example, the BioNLP 2024 "Discharge Me!" [90] challenge focuses on the generation of Brief Hospital Course and Discharge Instructions from emergency department records.

**Automated document generation**

Automated document generation refers to the generation of medical documents from various data sources, including structured clinical variables, unstructured free-text, or multimodal inputs. These systems aim to reduce administrative burden, standardize note formats, and improve the completeness and consistency of documentation[91]. A long-standing task is to translate discrete clinical variables—such as ICD codes, laboratory values, medication lists, or vital signs—into clinical narratives. For example, J Kurisinkel & Chen[34] applied NLG to transform patient admission ICD codes into discharge instructions. Soni & Dina [92] used structured and tabular information in EHR to automatically generate progress notes.

Many studies have focused on **Medical Report Generation** from medical images and multiple modalities data sources such as chest X-rays, CT scans, ECG signals, or histopathology slides. Yang et al.[93] applied sequence-to-sequence models to generate medical reports from ECG data. Many studies explored radiology report generation to reduce the burden of radiologists[94,95]. Chest X-ray images are widely explored through the release of public datasets such as MIMIC-CXR[96] and IU X-Ray[97]. There are a few studies, e.g., Paalvast et al.[69], that address other types of data. Other techniques and resources used in NLG include memory modules[72,98], knowledge graphs[99], and reinforcement learning[68,73]. There is an increasing interest in adding additional information to medical images, such as patient clinical history [100], knowledge base[101], and knowledge graph[99] to improve text generation.

**Data Augmentation**

Data augmentation has emerged as a critical application to account for the long-standing challenges of data scarcity, privacy restrictions[102,103], and annotation costs. By generating synthetic medical text that mimics the structure, content, and semantics of real-world medical records, researchers aim to create high-quality, privacy-preserving datasets for model training, benchmarking, and evaluation, without compromising patient privacy. [102,103].

For instance, Guan et al. [24] used respiratory disease features, such as “pneumonia” or “lung cancer,” to generate synthetic clinical note sections. Lee[25] proposed an RNN-based NLG model to generate synthetic chief complaints from structured input variables such as age, gender, and discharge diagnosis. Melamud & Shivade [33] proposed methods to generate privacy-preserved synthetic discharge summaries. Hoogi et al.[104] generated synthetic mammography reports to extend labeled medical report corpora. Amin-Nejad et al.[35] and Peng et al.[7] explored decoder-only generative LLMs to generate synthetic clinical data.

There is an increasing interest in using synthetic clinical narratives—such as chief complaints[25], discharge summaries[33], radiology reports[104], and medication histories—to augment training data for various applications, including text classification, readmission prediction, phenotype identification, and disease diagnosis support. For instance, synthetic data has been shown to improve cerebrovascular disease classification, where annotated data is typically limited[105]. Other studies have applied synthetic note generation to support few-shot learning, enabling NLP systems for low-resource settings with minimal real data[106]. Synthetic data generation has also been applied to probabilistic modeling of longitudinal EHRs, simulation of pharmacokinetic/pharmacodynamic (PK/PD) profiles, and the construction of virtual patients for use in drug development and population-level modeling[107]. These applications support data sharing across institutions, improve representation of underrepresented patient groups, and enable studies in data-constrained environments.

## Medical Dialogue Generation

Medical dialogue generation aims to provide clinically meaningful, context-aware, and empathetic responses in real-time interactions between healthcare providers and patients. These systems must handle evolving dialogue context, interpret patient intent, and deliver safe, accurate, and appropriate responses[108]. Recent studies[109,110] focused on multi-turn medical dialogue systems that closely mimic real-world doctor–patient interactions to reduce the burden on healthcare professionals. Dialogue generation has increasingly been integrated into intelligent consultation systems for early-stage disease diagnosis[111] and chronic care support. For instance, privacy-aware conversational agents have been developed to assist patients with chronic conditions like hypertension, offering tailored advice and interactive monitoring while preserving data confidentiality[112]. Recent studies[110,113,114] have explored the quality, factual grounding, and interpretability of dialogue responses, particularly in virtual care environments..

**Discussion**

This systematic review highlights the rapid development of NLG in the medical domain. Over the past 5 years, NLG has witnessed rapid progress from early-stage CNNs and RNNs to the dominant transformer-based solutions that are based on the attention mechanism, self-supervised training, and encoder-decoder architecture[115]. Multiple modalities of medical data have been utilized for NLG, increasingly impacting medical discovery and healthcare applications.

NLG is transforming clinical documentation, decision-making, and patient care through diverse and innovative uses[116]. One important contribution of NLG to healthcare is reducing the burden on healthcare providers. Clinicians constantly document, interpret, and summarize critical information in electronic health records (EHRs) to facilitate continuous and coordinated care. Appropriately summarizing patients' clinical care information is a critical skill for clinicians to communicate with patients (e.g., discharge summaries), handoff work at the change of shift, and work with colleagues with different expertise and across different clinical departments to handle complex cases (e.g., patients with multiple illnesses). However, the growing burden of clinical documentation has been a significant challenge in healthcare, contributing to physician burnout, reducing the efficiency of care delivery, and potentially compromising patient safety. Through automated summarization applications, NLG helps healthcare providers summarize various patient reports and doctor-patient conversations into concise summaries. Through automated documentation applications, NLG helps streamline the documentation process, reduce clinician burnout, and enhance healthcare delivery efficiency[117]. Through medical dialogue systems, NLG helps healthcare providers and patients interact by enhancing telemedicine consultations and mental health counseling[118].

The sensitive nature of clinical text causes barriers to sharing patients' data for research. Though automated de-identification systems that can be used to remove protected health information (PHI) exist, no systems can achieve 100% accuracy. Through synthetic medical text generation, NLG helps create large-scale, privacy-preserving synthetic datasets essential for medical research[119] and healthcare technology innovation[120]. Compared with the open domain, there are no large-scale datasets available to facilitate the large-scale training of medical LLMs. We have almost exhausted all the electronic data generated on this planet. Synthetic text generation is a promising solution to use AI to generate synthetic datasets to scale up datasets, similar to the AlphaZero system, which defeated human champions in the GO game, and was developed using synthetic records generated by AI without using any real-world human data.

Current NLG applications have limitations. One significant challenge is ensuring the safety and reliability of NLG to control errors that can lead to severe consequences for patient care[121,122]. Another limitation is the potential of generating biased or non-compliant text, as models trained on historical data may inadvertently perpetuate existing biases or fail to adhere to current medical guidelines and standards[123,124]. The evaluation of NLG systems needs to be improved. Most studies use automatic evaluation metrics such as ROUGE and BLEU scores, which are computationally efficient and allow for the rapid assessment of large volumes of generated content. However, automatic evaluation metrics cannot account for the language variations and nuanced quality of the generated texts that cannot be measured using the surface-form text. Human evaluation metrics, such as the Likert scale, allow evaluators to rate various aspects of the generated text. Yet, they are very expensive in demand for domain experts and hard to scale up. Future research should also examine the AI ethics, interpretability, and transparency of NLG applications, as it is crucial for healthcare providers to understand and trust NLG systems[125].

**Future work**

Future work should continue to explore multimodalities of medical data, such as clinical narratives, medical images, and omics data. In addition, exploring the development of voice-activated NLG systems to enable hands-free operation, which can be particularly useful in clinical settings such as surgical operations. For example, the use of vision transformers to process medical images in conjunction with text generation tasks. Human-in-the-loop is also an important topic for future studies. As no AI systems can be 100% accurate, it is important to enable clinicians to provide feedback for NLG systems to help refine the models and improve the quality and relevance of the generated content. Developing evaluation metrics that go beyond traditional surface measures like ROUGE and BLEU could help the advancement of NLG. The evaluation metrics should account for diverse aspects beyond the surface form of text, such as clinical relevance, factual accuracy, and potential impact on patient outcomes. Future work should also explore the efficient integration of NLG systems into real-world clinical workflows and test the efficacy of AI implementation.

## Limitation

This study has limitations as we focus on NLG methods and applications, studies that about the potential bias, security, risks, interpretability, and AI ethics, but without concrete applications were not included. The exclusion of non-English studies potentially narrows the scope of review, especially for NLG in low-resource languages.

## Methods

### Data Sources and Search Strategy

We searched across six academic databases, including PubMed, ACM Digital Library, Web of Science (WoS), Science Direct, Scopus, and Embase, for peer-reviewed articles. To capture relevant conference proceedings, the Association for Computational Linguistics (ACL)

Anthology was included. The search targeted peer-reviewed articles published between January 1, 2018, and December 31, 2024. The search used grouped keywords related to natural language generation (e.g., "text generation," "natural language generation," "sequence-to-sequence") and healthcare (e.g., "clinical," "medicine," "EHR," "EMR"). As shown in **Table 3**. Only articles published in English were included.

**Table 3.** Literature search using keywords to identify articles**.**

| Search | | |
|---|---|---|
| Search Strings | Text Generation Group | 1. Text Generation; Report Generation |
| | | 2. Natural Language Generation; NLG |
| | | 3. Sequence to Sequence; Seq to Seq |
| | Healthcare Group | 4. Clinical; Medicine; Healthcare; Health |
| | | 5. EHR; EMR; Electronic Medical Records; Electronic Health Records |
| Combinations | | 6. 1 OR 2 OR 3 |
| | | 7. 4 OR 5 |
| | | 8. 6 AND 7 |

**Eligibility Criteria, source, and search**

We included studies that met the following criteria: (1) published as peer-reviewed original research articles between January 1, 2018, and December 31, 2024, and (2) presented concrete NLG methods or applications applicable to healthcare or medical research. Studies were excluded if they were editorials, commentaries, reviews, or focused solely on non-clinical applications (e.g., social media communication). Articles without full texts available were excluded.

We conducted a two-stage screening protocol using Covidence, a web-based systematic literature review tool. Duplicates were automatically removed by Covidence. Database searches were conducted in January 2025. Six reviewers screened titles and abstracts, with each article randomly assigned to two reviewers. Disagreements were resolved by a third reviewer. Articles that passed the title and abstract review were screened in the second-stage full-text review. To ensure good agreements among all reviewers, three training sessions were conducted. In each session, all reviewers reviewed a shared set of 15 articles, followed by group discussions to resolve discrepancies and refine the review guidelines.

### Data extraction and analysis

Data extraction was performed by six reviewers using a standardized spreadsheet. Studies that passed the two-stage screening were reviewed by at least two reviewers, who also extracted predefined information in the review guidelines. Discrepancies between reviewers were resolved through group discussion and consensus. We determined a total of 10 data elements to extract, including: NLG category (e.g., text-to-text, image-to-text), model architecture (e.g., decoder-based, GAN), encoder architecture (indicate the architecture of the encoder module, e.g., LSTM), decoder architecture (indicate the architecture of the decoder module, e.g., GPT), data access (i.e., private or public), data source, clinical application, evaluation methods (i.e., machine evaluation or human evaluation), machine evaluation metrics (e.g., BLEU, ROUGE, METEOR), human evaluation metrics (e.g., Likert scale ratings, expert assessments).

Following data extraction, studies were grouped by clinical application categories to enable thematic synthesis. For each application area, we analyzed trends in model architecture, data types and sources, evaluation strategies, and performance metrics.

**Acknowledgments**

**Funding**: This study was partially supported by grants from the Patient-Centered Outcomes Research Institute® (PCORI®) Award (ME-2018C3-14754, ME-2023C3-35934), the PARADIGM program awarded by the Advanced Research Projects Agency for Health (ARPA-H), National Institute on Aging, NIA R56AG069880, National Institute of Allergy and Infectious Diseases, NIAID R01AI172875, National Heart, Lung, and Blood Institute, R01HL169277, National Institute on Drug Abuse, NIDA R01DA050676, R01DA057886, the National Cancer Institute, NCI R37CA272473, and the UF Clinical and Translational Science Institute. The content is solely the responsibility of the authors and does not necessarily represent the official views of the funding institutions. We gratefully acknowledge the support of NVIDIA Corporation and the NVIDIA AI Technology Center (NVAITC) UF program.

**Author contributions**

ML and YW was responsible for the overall design, development, and evaluation of this study. ML, XH, ZY, JP, CP, ST had full access to all the data in the study and takes responsibility for the integrity of the data and the accuracy of the data analysis. ML and YW did the bulk of the writing; XH, ZY, CP, JP, and ST contributed to the writing and editing of this manuscript. All authors reviewed the manuscript critically for scientific content, and all authors gave final approval of the manuscript for publication.

**Competing interests**

The Authors declare no Competing Financial or Non-Financial Interests.

**Materials & Correspondence**: Yonghui Wu, PhD

**Statistical information: N/A**